\documentclass{article}

\usepackage{PRIMEarxiv}

\usepackage[utf8]{inputenc} 
\usepackage[T1]{fontenc}    
\usepackage{hyperref}       
\usepackage{url}            
\usepackage{booktabs}       
\usepackage{amsfonts}       
\usepackage{nicefrac}       
\usepackage{microtype}      
\usepackage{lipsum}
\usepackage{fancyhdr}       
\usepackage{graphicx}       
\graphicspath{{media/}}     
\usepackage{amsmath,amssymb,amsfonts}
\usepackage{algorithmic}
\usepackage{graphicx}
\usepackage{textcomp}
\usepackage{xcolor}
\usepackage{multirow}
\usepackage{verbatim}
\usepackage{arydshln}
\usepackage{hhline}

\title{Dimension Reduction for time series with Variational AutoEncoders}

\author{
  William Todo $^\dag \: ^{*}$, B\'eatrice  Laurent {$^\dag$}, Jean-Michel Loubes {$^\dag$}, Merwann Selmani $^*$ \\
 {$^\dag$} Institut de Math\'ematiques de Toulouse \\
  University of Toulouse  \\
*  Liebherr Aerospace Toulouse \\
  University Toulouse 3 \\
}

\begin{document}
\maketitle

\begin{abstract}
 In this work, we explore dimensionality reduction techniques for univariate and multivariate time series data. We especially conduct a comparison between wavelet decomposition and convolutional variational autoencoders for dimension reduction. We show that variational autoencoders are a good option for reducing the dimension of high dimensional data like ECG. We make these comparisons on a real world, publicly available, ECG dataset that has lots of variability and use the reconstruction error as the metric. We then explore the robustness of these models with noisy data whether for training or inference. These tests are intended to reflect the problems that exist in real-world time series data and the VAE was robust to both tests.

\end{abstract}


\section{Context}
The curse of dimensionality is at the heart of decades of research in statistics and machine learning, preventing the use of a large number of methods when the dimension of the data increases. Yet with the exponential growth of data collection, high dimensional data such as images, time series or functional data have become more and more studied.  Traditional Machine Learning techniques are not a good fit for this kind of data first due to the practical difficulty to handle from a computational point of view this kind of database but also from a theoretical point of view since the accuracy of the algorithms hamper for high dimensional data.  To overcome theses issues, one well studied way of preprocessing the data is to reduce its dimensionality. Principal Components Analysis (PCA) \cite{PCA} is a well known general technique that has been widely studied together with  Independent Components Analysis (ICA) \cite{ICA} or Factor Analysis (FA) \cite{FA} and more. Recently, with the emergence of AutoEncoders (AE) and Variational AutoEncoders (VAE) \cite{kingmavae} more complex data driven dimensionality reduction techniques became possible and are benchmarked (\cite{VAE_bench1} $\&$ \cite{VAE_bench2} ).

 Time Series and Multivariate Time Series require a special attention. Actually  classical dimensionality reduction techniques fail to capture the temporal aspect of the observations, therefore there are some specific techniques for that kind of data, in particular projection methods onto specific bases for instance onto generical wavelet basis as in~\cite{wav} or data driven basis using for instance Functional Principal Component Analysis (FPCA) \cite{fpca}. Wavelet transform uses a low-pass filter to extract low frequency information and a high-pass filter to extract high frequency information. The advantage over the Fourier Transform is that the positional information is conserved with the wavelet transform. FPCA decomposes functional data into basis functions that explain the variance.  The outcome of such method is to discover features  using such projections, that are expected to concentrate the information on a small number of coefficients. Dimension reduction is key to handle the information contained in the time series and is often used in clustering \cite{javed2020benchmark} or anomaly detection \cite{barreyre2019multiple} for example.\\
 Very recently, Machine Learning methods using deep neural networks have been considered as alternatives to such methods. They enable to construct low dimension embedding by considering  the features from the penultimate neural layer that are used to build the forecast. In the same vein, variational autoencoders map the data into a structured representation of lower dimension in a data driven way. However, the use of variational autoencoders on time series as a dimension reduction technique is not yet well studied or compared to other methodologies.

This paper shows the advantages of VAE regarding the dimensionality reduction power and robustness against more traditional methods like Wavelet decomposition and FPCA on real world ECG datasets.


In this paper, we trained various convolutional variational autoencoders to obtain  a lower  dimensional representation  of the dataset PTB-XL \cite{ptbxldataset}. We compare those results with the state of the art method of the wavelet decomposition. Then we stress the tests to highlight the good properties of VAEs. We also test these VAEs on other real world ECG datasets from the Physionet challenge \cite{physionet_challenge}. The code and models for these experiments are be available on github.

\begin{figure}
    \centering
    \includegraphics[width=\linewidth]{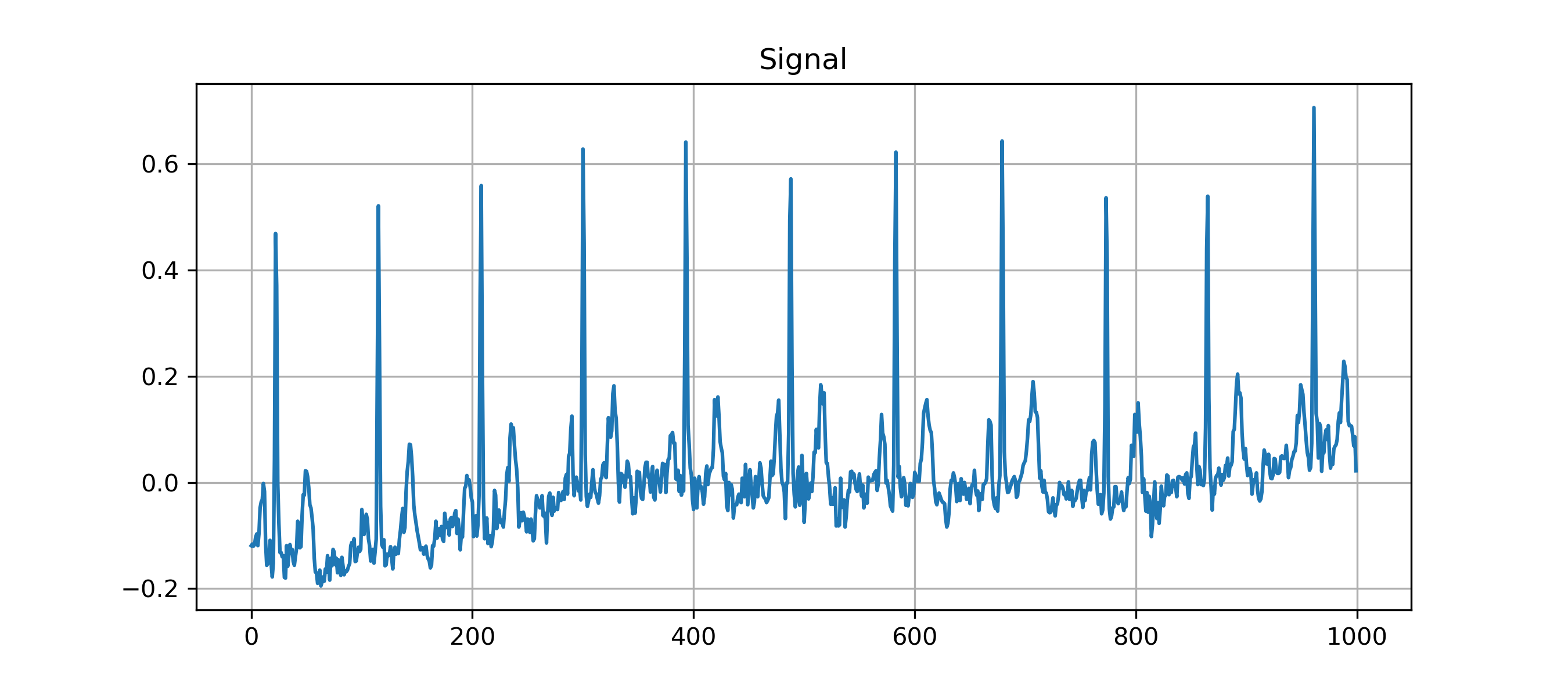}
    \caption{Example signal from the dataset PTB-XL}
    \label{fig:wavelet_decomposition_a}
\end{figure}

For this study, we use a real world dataset of ECG measurements : PTB-XL dataset. On this dataset each sample corresponds to 12 recordings of electrical activity on the body surface during 10 seconds at a frequency of 100 Hz. Each sample is a 12000 dimensional point, hence the need to reduce the dimensionality. For the sake of the study, we use also a cropped version of that dataset for univariate experiments. The cropped version consists of all the sample, but with only the first ECG so each sample of that version is composed of 1000 points as shown in Fig. \ref{fig:wavelet_decomposition_a}.

\section{One dimensional time series}

\subsection{Method}
In this section, we describe the architecture of the VAEs, the training process and some notes about the implementation.We use Pytorch \cite{pytorch} to implement the VAE since this well-known library allows for a lot of flexibility to implement these objects.

\subsubsection{Variational auto encoder setup}
Before settling for a convolutional variational autoencoder we also tested RNN based VAEs and, although the performance was similar, the training time was longer. 
That is why experiments were done using the CNN VAE. The encoder and decoder are symmetrical with a 3-layers deep convolutional neural net. The number of convolutional filters 
is a parameters that we change depending on the dimension of the latent space to have the best MAE score possible. 
The training is done using 256-length crop of the signal. It allows us to add data augmentation similar to what is done in \cite{ptbxlbenchmark} in the training process.
We randomly crop the input signal from 1000 data points to 256 data points. These crops help the model to better generalise because regions of interests in the signal are not always in the same area and heartbeat are not synchronised. Because the heart rate is not stable throughout the dataset, we artificially change the cardiac rhythm by up-sampling and down-sampling the input. We take that crop size into account when computing the compression rates. We also compute the reconstruction error as the mean of four VAE reconstruction errors so the error is computed with the whole signal and not just a cropped one.

\subsubsection{Wavelet transformation} 
\label{subsection:wavelet_transformation}
\begin{figure}
    \centering
    \includegraphics[width=\linewidth]{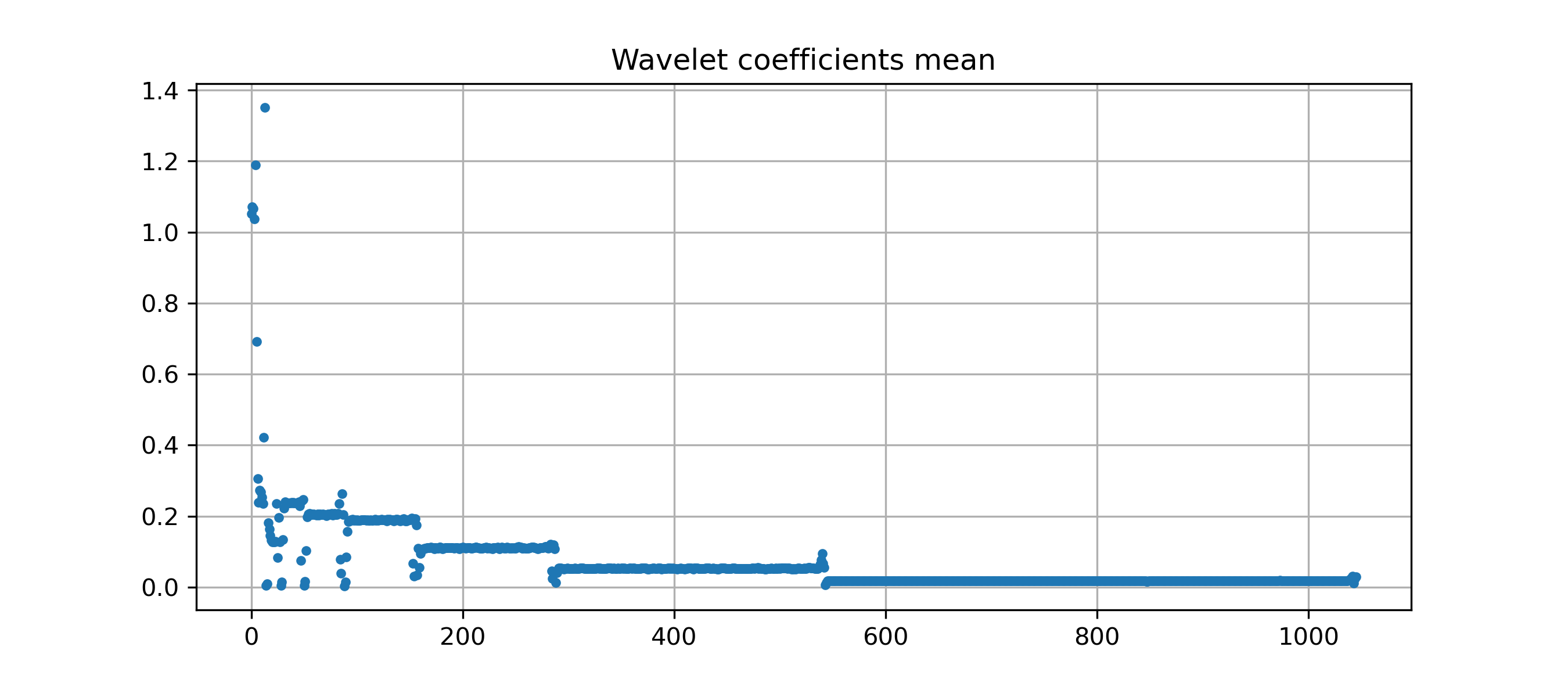}
    \caption{Mean throughout the dataset of the energy of the wavelet decomposition}
    \label{fig:wavelet_decomposition_c}
\end{figure}

We compare the VAE compression with wavelet decomposition technique \cite{waveletcompression}. Wavelet transform concentrates the signal information into a small number of coefficients. It is possible to perform lossy compression, eliminating coefficients with small magnitude.
This concentration of information is powerfull and is still used today including in the JPEG 2000  image compression standard \cite{JPEG2000}.
Our compression rate is computed counting the kept coefficients. We compress our ECG signals using two different methods: 
The first one is the one we should use in a real world setting: we compute the wavelet transform for all the dataset and then we keep the $n$ coefficients with the highest mean (in absolute values), this is the Global Approach. Because this dataset presents non synchronous time series, the localisation of features of interest are not fixed. The mean energy of the wavelet coefficients throughout the dataset does not reflect the localisation of a feature but the level of decomposition. In the wavelet decomposition the first wavelet coefficients are those coming from the highest level of decomposition and thus of energy. That's why this method
is quite similar to keeping the $n$ first wavelet coefficients we show that in Fig. \ref{fig:wavelet_decomposition_c}.

%

The other method consists in keeping the $n$ biggest coefficients per signal. Although this method performs much better than the first one it is unrealistic in practice because we need to know which coefficient we kept in order to reconstruct the signal. This is an oracle showing the best case scenario for the wavelet decomposition.

\subsubsection{FPCA on ECG data}
A common way to reduce dimensionality on data is to use principal components analysis \cite{PCA}. Since our data are Time series it is natural here to think of Functional Principal Component Analysis \cite{fpca}. This technique works well with certain kind of data as shown in \cite{FPCAcompression} and has the advantage to offer lots of insights on the data. As told before our dataset is not synchronous and with very diverse time series shapes that reduce our ability to extract functional bases that represent well the data. This technique, although successful in a wide range of functional datasets is not well suited for ECG data.

\subsection{Reconstruction}
\begin{figure}
    \centering
    \includegraphics[width=\linewidth]{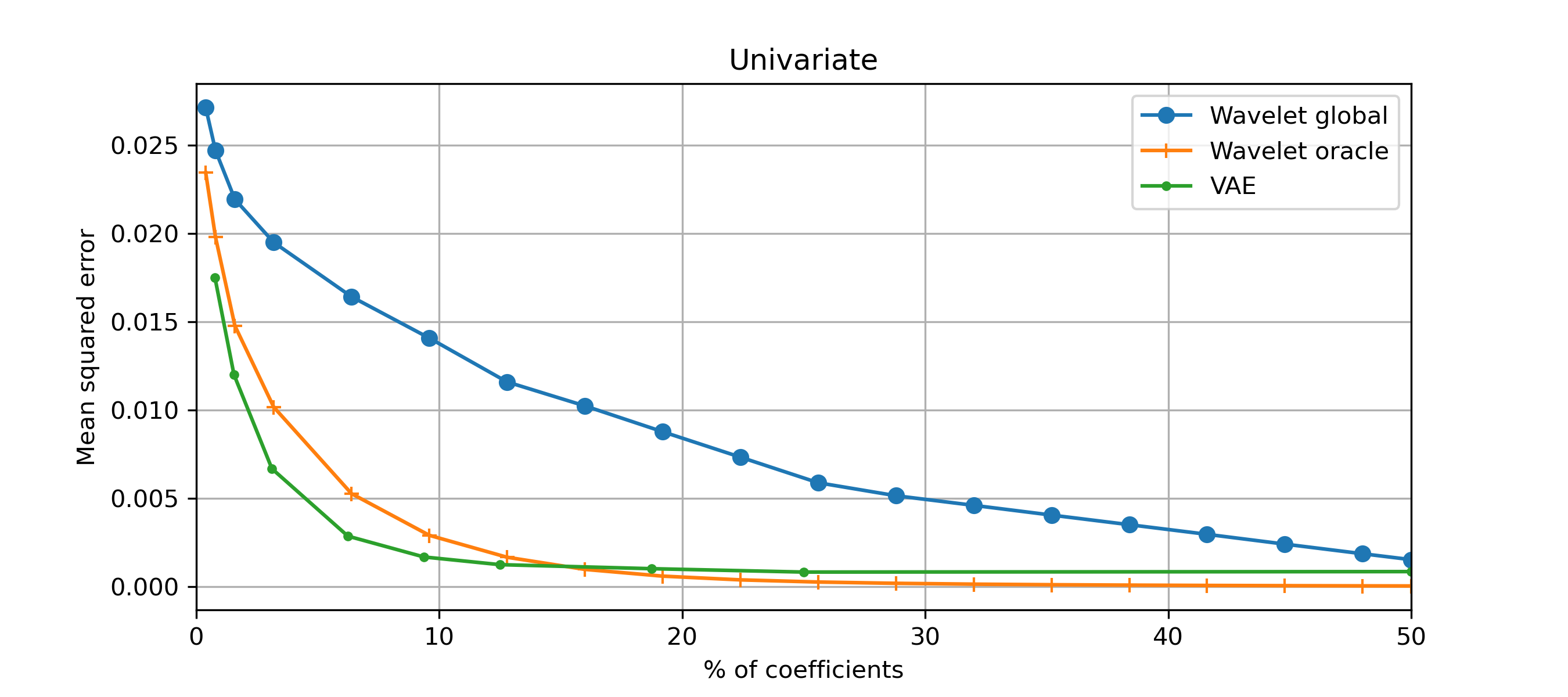}
    \caption{Performance comparison between VAE and wavelets giving the percentage of kept coefficients}
    \label{fig:univariate_perf}
\end{figure}

In this section we try different methods of dimensionality reduction and compare them according to the reconstruction error. We use the mean squared error between the original and the reconstructed signal as our metric. We compare VAE performance against the two wavelet decomposition approaches in Fig. \ref{fig:univariate_perf}. 

Variational autoencoders outperform both the standard wavelet decomposition method, but also the oracle when it comes to reduce the data to a really low dimension (compression rate of 10 and more). This was expected because the tailored convolutional filters of the VAE are a better fit than non data driven wavelet basis. The VAE has a better reconstruction error than the wavelet decomposition. Even with a lower compression rate it is still not far from the performance achieved by the unrealistic oracle.
Variational autoencoders present lots of advantages against wavelet transform in a high compression rate scenario first because of a lower error rate and second because of its well structured latent space that can easily be used for further processing. Although an important and interesting topic, the feature extraction capabilities of variational autoencoders are not discussed here.

\section{Multidimensional time series}

\subsection{Reconstruction}
\label{subsection:multidim_reconstruction}
Here, as explained previously in Section \ref{subsection:wavelet_transformation}, we keep the $n$ largest coefficients from the decomposed signals. This means that we will not necessarily keep the same number of coefficients for each dimension and that it is even possible to completely ignore one dimension if there are just a few coefficients kept.

\begin{figure}
    \centering
    \includegraphics[width=\linewidth]{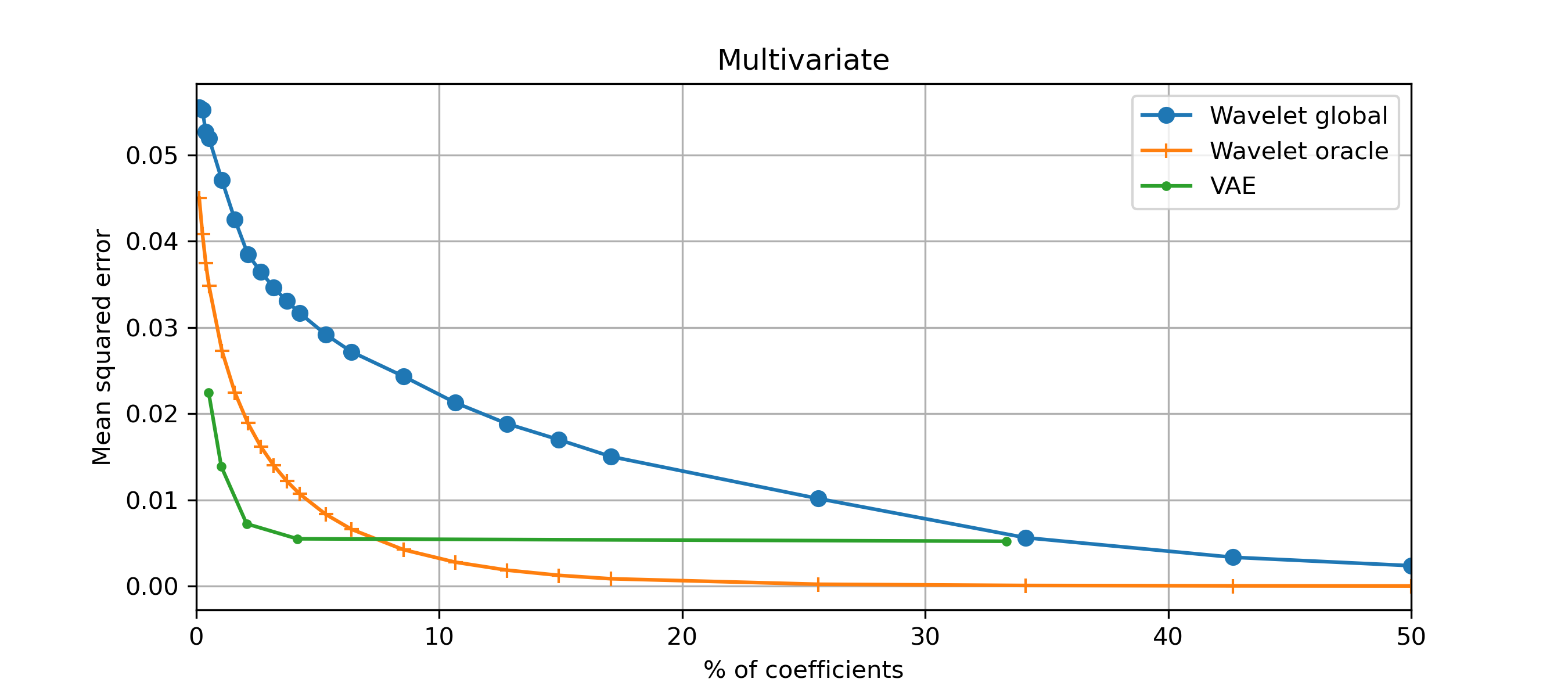}
    \caption{Performance comparison between VAE and wavelets giving the percentage of kept coefficients on PTBXL dataset}
    \label{fig:multivariate_perf}
\end{figure}

CNN VAE uses filters that can combine information from all dimensions of the time series and the 12 Lead ECG data are very correlated because they all see the same heartbeat from different positions. That is why the CNN VAE should be able to perform a really good dimensional reduction on these data. In practice, it is true when we keep a tiny amount of coefficients to reconstruct the data. We can see that phenomenon in Fig. \ref{fig:multivariate_perf}. When we retain less than $6\%$ of the coefficients, the reconstruction error is even much better than the wavelet oracle. After, the model does not really improve despite the increase in the size of the latent space. We could explain that with the difficulty to untangle the information contained in the latent space. 

To highlight the generalization capability we test the VAE trained on the PTBXL dataset on two other datasets from the physionet challenge \cite{physionet_challenge} : $10,344$ recordings from Georgia, named "Georgia" and $3,453$ recordings of unused data from CPSC2018 \cite{physio_china} named China.
We use the preprocessing from \cite{physio_preprocesing}, adding a scaling operation and resampling to 100Hz to match PTBXL.
The results are summarized in Table \ref{tab:data_com} where the numbers are bolded when the VAE results are better than the global wavelet method and underlined when the VAE is better than the wavelet oracle. We observe the same trends on all three data sets: VAEs are also better than the wavelet method for high compression ratio, even compared to the oracle. 

\begin{table}
\centering
\caption{Dataset Comparison}
\label{tab:data_com}
\begin{tabular}{c|l|lll}
Dataset                  & \multicolumn{1}{c|}{Method} & \multicolumn{3}{c}{Mean Squared Error}                                         \\ 
\hline
\multirow{3}{*}{PTB}     & VAE                         & \textbf{\underline{0.02201}} & \textbf{\underline{0.00723}} & \textbf{0.00487}         \\ 
\cdashline{2-2}[1pt/1pt]
                         & Global                      & 0.05074                  & 0.03815                  & 0.00603                  \\ 
\cdashline{2-2}[1pt/1pt]
                         & Oracle                      & 0.03624                  & 0.02002                  & 0.00008                  \\ 
\hline
\multirow{3}{*}{Georgia} & VAE                         & \textbf{\underline{0.01759}} & \textbf{\underline{0.00723}} & \textbf{0.00531}         \\ 
\cdashline{2-2}[1pt/1pt]
                         & Global                      & 0.04057                  & 0.03220                  & 0.00576                  \\ 
\cdashline{2-2}[1pt/1pt]
                         & Oracle                      & 0.02871                  & 0.01612                  & 0.00008                  \\ 
\hline
\multirow{3}{*}{China}   & VAE                         & \textbf{\underline{0.03632}} & \textbf{\underline{0.01972}} & 0.01424                  \\ 
\cdashline{2-2}[1pt/1pt]
                         & Global                      & 0.06356                  & 0.05313                  & 0.00710                  \\ 
\cdashline{2-2}[1pt/1pt]
                         & Oracle                      & 0.04053                  & 0.02175                  & 0.00009                  \\ 
\hline
\multicolumn{2}{c|}{~Coefficients $\%$}                     & \multicolumn{1}{c}{0.5}  & \multicolumn{1}{c}{2}    & \multicolumn{1}{c}{33 } 
\end{tabular}
\end{table}


\begin{figure}
    \centering
    \includegraphics[width=\linewidth]{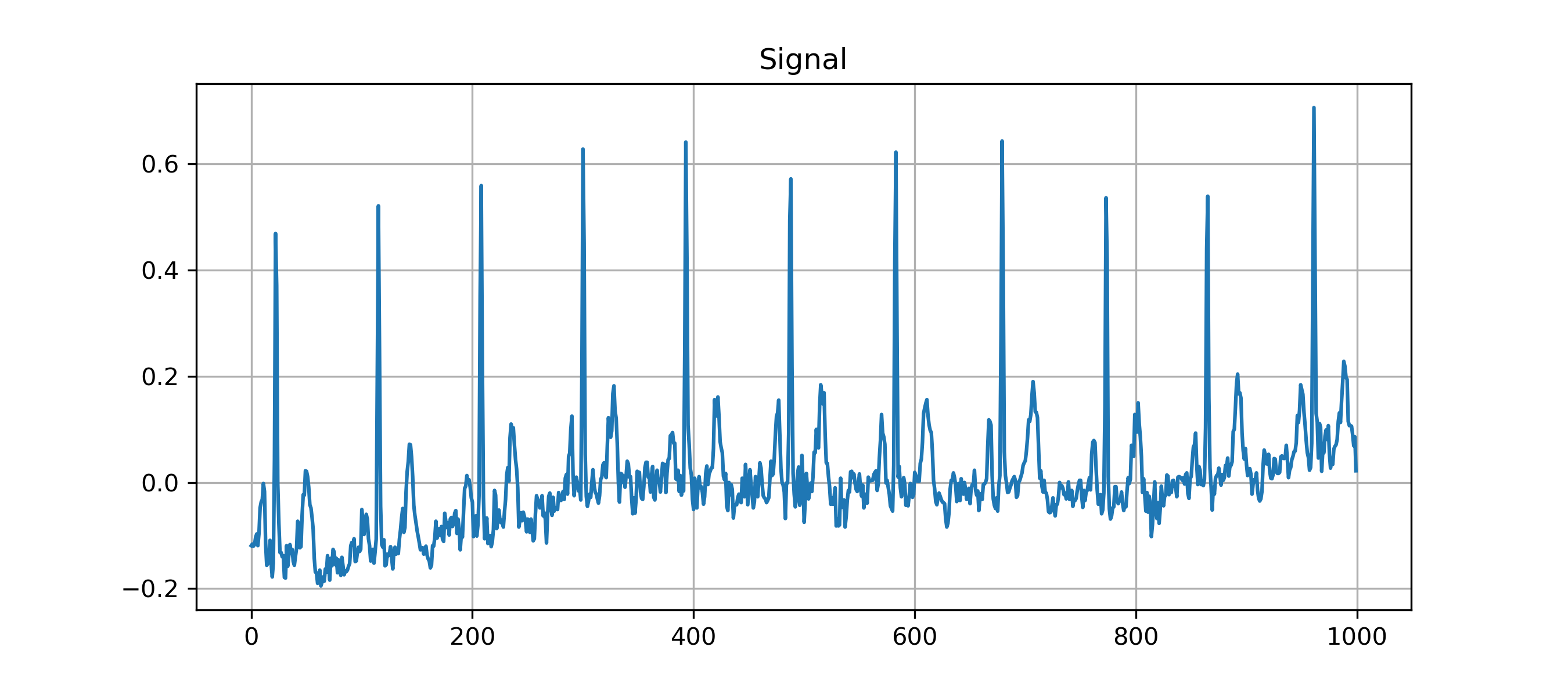}
    \caption{Original ECG without noise}
    \label{fig:noise_original}
\end{figure}

\begin{figure}
    \centering
    \includegraphics[width=\linewidth]{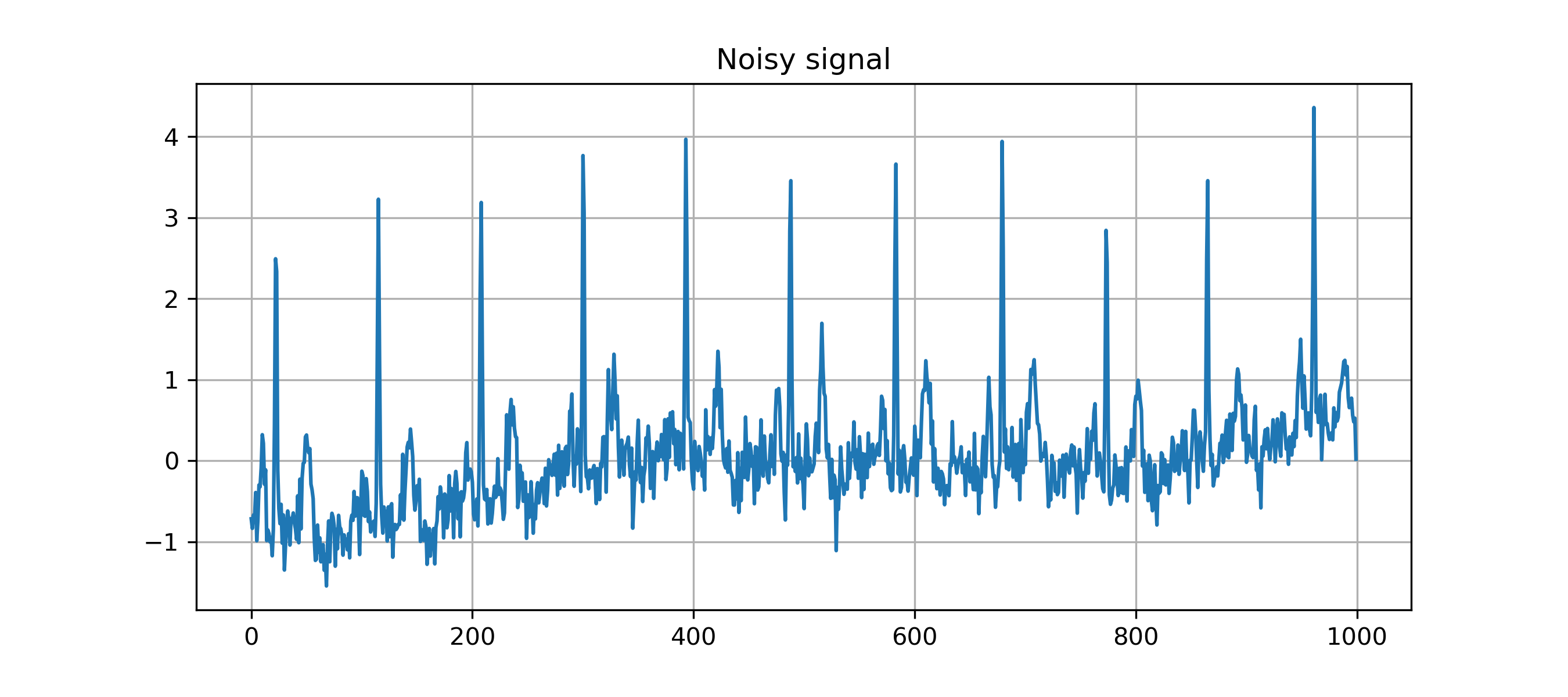}
    \caption{ECG with added noise}
    \label{fig:noise_noisy}
\end{figure}

In this section we present the experiments done to assess the robustness of VAE feature extractions on noisy data because wavelet decomposition is often used as a denoising tool. Eliminating small coefficients in the wavelet decomposition using hard or soft thresholds tends to eliminate noise in the reconstructed signal.(\cite{wavelet_denoising_0}, \cite{wavelet_denoising} ). We add white noise to our time series, it is illustrated in Fig. \ref{fig:noise_noisy} and represent $20\%$ of the signal variance. This is done in two ways: first we place ourselves in a noisy environment to train the VAE, because in a real environment, the collected data can be noisy. During testing, we take a noisy time series and we compare its reconstruction with the original time series to see how the VAE is affected by the noise. The other aspect we study is the robustness of the VAE to noise during inference. This involves a VAE trained with low-noise data which encounters noisy data during inference.

\begin{figure}
    \centering
    \includegraphics[width=\linewidth]{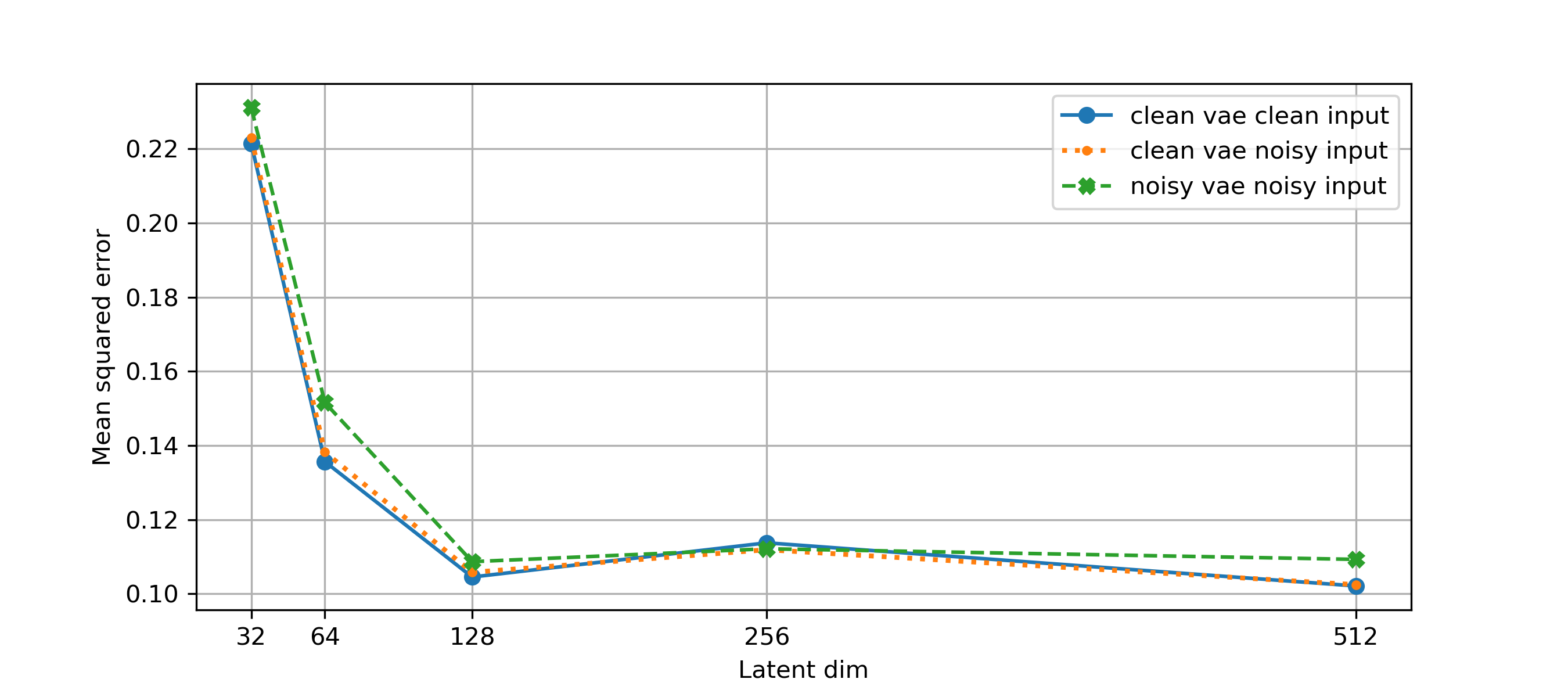}
    \caption{Performance comparison for VAE trained on various kinds of data}
    \label{fig:noise_stdy}
\end{figure}

We can see in Fig. \ref{fig:noise_stdy} that a VAE trained on clean data and tested on noisy data (robustness experiment) performs well and there is just a small difference with the VAE tested on clean data. That experiment shows the robustness to noise of the VAE. Also, this figure shows us that the VAE trained on noisy data is still performing quite well, even if there is some loss of information.

\subsection{Anomaly detection}

The specificity of the VAE used for dimension reduction is that  it learns specific convolutional filters adapted to the data, contrary to the wavelet decomposition which is not data driven. As shown in Section \ref{subsection:multidim_reconstruction} this proves useful for keeping reconstruction error relatively low, even with a high compression ratio.

Yet another field of common application of VAE is anomaly detection. Hereafter, we test if the VAE trained for a  dimension reduction purpose can also be used for anomaly detection, without changing the training process.

Actually, we train a VAE only on data without a detected disease. We do not do any tuning on that architecture so the parameters and hyperparameters are the same as in the other VAE. We use the simplest method to make the predictions by considering that a bad reconstruction rate indicates an anomaly. For this, set a threshold on the reconstruction error and consider the forecast rule that a reconstruction error  greater than that threshold is used to predict an anomaly. The results are presented in the confusion matrix Table \ref{tab:cm_wav} and Table \ref{tab:cm_vae}. It is better than the results using the wavelet decomposition because of its lack of adaptability.
$69\%$ of accuracy on that dataset is not that high, but it still shows the adaptability of VAE on the training dataset.


\begin{table}
\begin{tabular}{l|l|c|c|c}
\multicolumn{2}{c}{}&\multicolumn{2}{c}{Predicted label}\\
\cline{3-4}
\multicolumn{2}{c|}{}&Anomaly&Normal\\
\cline{2-4}
\multirow{2}{*}{True Label}& Anomaly & $55\%$ & $45\%$ \\
\cline{2-4}
& Normal & $45\%$ & $55\%$  \\
\cline{2-4}
\end{tabular}

\caption{\label{tab:cm_wav} Confusion matrix for anomaly detection using wavelets reconstruction errors}
\end{table}
\vskip .1in
\begin{table}
\begin{tabular}{l|l|c|c|c}
\multicolumn{2}{c}{}&\multicolumn{2}{c}{Predicted label}\\
\cline{3-4}
\multicolumn{2}{c|}{}&Anomaly&Normal\\
\cline{2-4}
\multirow{2}{*}{True Label}& Anomaly & $69\%$ & $31\%$ \\
\cline{2-4}
& Normal & $31\%$ & $69\%$  \\
\cline{2-4}
\end{tabular}
\caption{\label{tab:cm_vae} Confusion matrix for anomaly detection using VAE reconstruction errors}
\end{table}
\section{Conclusion}
\label{section:conclusion}
In this work we compare data driven dimension reduction techniques such as Functional PCA, LSTM-VAE and CNN-VAE to classical wavelet decomposition on a real world ECG dataset. We selected the CNN-VAE as the most suitable architecture for this specific problem.
In addition to having better performance, especially when the dimension is reduced to small spaces, we show that it is robust to noise both during training and during inference and that it also has some anomaly detection capabilities.

In future work, it would be interesting to explore the issue of disentanglement of variational autoencoders latent space to further improve the performance of this multivariate time series dimension reduction.

\section*{Acknowledgments}
The work is  supported by the AI Interdisciplinary Institute ANITI, which is funded by the French 'Investing
for the Future – PIA3' program under the Grant agreement ANR-19-PI3A-0004.

\end{document}